\documentclass[conference]{IEEEtran}
\IEEEoverridecommandlockouts

\usepackage[T1]{fontenc}
\usepackage[utf8]{inputenc}
\usepackage{newtxtext,newtxmath}
\usepackage{graphicx}
\usepackage{booktabs}
\usepackage{amsmath}
\usepackage{array}
\usepackage{stfloats}
\usepackage{url}
\usepackage[hidelinks]{hyperref}

\newcommand{\imagedir}{images}

\begin{document}

\title{CoupVisor: Strategy Optimization by Round\\ and Challenge Decision Support}

\author{\IEEEauthorblockN{Cris Huynh}
\IEEEauthorblockA{\textit{crishuynh2004@gmail.com}}}

\maketitle

\begin{abstract}
This paper presents CoupVisor, a decision-support system for the
hidden-information card game Coup. It addresses two questions: what a player
should do on each turn, and when a player should challenge an opponent's claim.
The system is built around a single description of game events, which is shared
across manual play, replay of recorded games, simulation, belief tracking,
advisor recommendations, and learning-based policies. CoupVisor estimates the
chance that a claim is truthful by combining how likely each role is with how
many cards the claimant still holds, which corrects a case where the very first
claim of a game was flagged as suspicious despite no evidence. We compare a
rule-following advisor and several learned and heuristic players across many
simulated games and different opponent styles. Our main finding is that the
choice of reward, whether it rewards short-term gains or ultimately winning the
game, decides which learning approach performs best, and that a win-oriented
reward produces a policy that outperforms all baselines.
\end{abstract}


\section{Introduction}
Coup is a card game in which every player holds two face-down role cards and may
claim any role at any time, whether or not the claim is true. Claims are enforced
only by challenges: a player who doubts a claim may demand proof, and whoever is
wrong about the claim loses a card. Every turn therefore contains two coupled
decisions, which action to take and whether to challenge the action someone else
has just claimed, and both must be made without seeing any opponent's hand. Coup
is thus a compact instance of a general problem: acting under uncertainty against
an adversary who benefits from being misread. A full description of the rules is
given in Appendix~\ref{app:game}.

What makes this problem tractable is that the evidence is already public. Every
action, block, challenge, and reveal is visible to the whole table, and each one
constrains which roles a player can plausibly hold. A player who claims Duke and
later claims Assassin has committed to a specific two-card hand. A role whose
three copies are all face up cannot be claimed truthfully by anyone. Yet at a
real table this record is almost never used. Players challenge on instinct, on
table talk, and on a vague sense that someone has claimed too much, and after the
game there is no way to say whether a given challenge was justified by the
information available at the time. The gap is not a shortage of evidence but the
absence of anything that turns the evidence into a number.

CoupVisor closes that gap. It is an observer and advisor rather than a
game-playing bot: it watches the public record, maintains an explicit probability
distribution over every opponent's hidden roles, and converts that distribution
into a stated answer to the question a player is actually asking, namely whether
this particular claim is worth challenging. The design decision that makes the
rest of the system possible is that a single event schema describes the public
record, and every component consumes it. Manual turn entry, replay of a recorded
game, the synthetic game simulator, the Bayesian belief tracker, the threshold
advisor, and the offline-trained policies all read the same events, so a
recommendation produced during live play, during replay, and during a training
run is produced by identical code.

This paper reports what that system does and how well it does it. We describe the
belief model and its evidence weights, the challenge rule and the thresholds it
applies, and the conversion from a per-card probability to the per-hand
probability that a claim actually asserts. We then evaluate the advisor over 500
simulated games spanning five seeds and two opponent conditions, ablate the belief
mode and the advisor itself, and benchmark behavior cloning, deep Q-learning,
three non-learning heuristics, and a random policy under two different reward
definitions.

The work pays off in three ways. For a player, an instinct becomes a stated
probability with a stated reason, separated cleanly from the hard rule
constraints such as the forced Coup at ten coins that no probability can
override. For research, the frozen evaluation gives a testbed in which belief
modelling, decision rules, and learned policies can each be varied on their own.
For decision support more generally, one result transfers beyond the game: under
a shaped short-term reward, behavior cloning beats deep Q-learning by a wide
margin, while under a reward aligned with actually winning, the same deep
Q-learning architecture beats every baseline. Nothing about the network or the
data changed between those two runs. Reward alignment, not algorithm choice,
decided the outcome.

\section{Background}
This section defines the technical terms used throughout the paper so that the
later sections can be read without external reference.

\subsection{Coup and the Public Record}
Coup uses five roles, Duke, Assassin, Captain, Ambassador, and Contessa, with
three copies of each in the deck. Each player begins with two hidden role cards,
called \emph{influence}, and two coins. Losing a challenge or being couped costs
one influence, and a player at zero influence is eliminated. Actions carry coin
costs and legality constraints: Assassinate requires three coins, Coup requires
seven, and a player holding ten or more coins \emph{must} Coup. A \emph{claim} is
the assertion of a role implied by an action, for example Tax implies Duke, or by
a block, for example blocking Assassinate implies Contessa. A \emph{challenge} is
a demand that the claim be proven by revealing the card. The sequence of actions,
claims, blocks, challenges, and reveals is visible to everyone and forms the
\emph{public record} on which CoupVisor operates.

\subsection{Imperfect Information and the POMDP}
A game is one of \emph{imperfect information} when a player cannot observe the
full state, here the opponents' hidden cards. The standard formalism is the
\emph{Partially Observable Markov Decision Process} (POMDP), defined by a set of
hidden states, a set of actions, a transition rule, a set of observations, and an
observation rule that links hidden states to what is actually seen. Because the
true state is unavailable, a POMDP agent acts on a \emph{belief state} instead.

\subsection{Belief, Prior, Posterior, and the Bayesian Update}
The \emph{belief} $b_i(r)$ is the probability that player $i$ holds role $r$
given the public record $h_t$ up to time $t$:
\begin{equation}
b_i(r) = P(r_i = r \mid h_t).
\end{equation}
The \emph{prior} is the belief before an event is observed; at the start of a
game it is uniform, $b_i(r) = 0.20$ for five roles. The \emph{posterior} is the
belief after the event. The two are linked by \emph{Bayes' rule}, which in the
proportional form used here states that the posterior is the prior multiplied by
the \emph{likelihood} of the observed event under each hypothesis:
\begin{equation}
b_i(r) \;\propto\; P(\mathrm{event} \mid r_i = r)\cdot b_i(r),
\label{eq:bayes}
\end{equation}
followed by \emph{normalisation}, that is, rescaling so the values sum to one
across roles.

The likelihood term is implemented as a \emph{likelihood ratio}, the ratio of how
often a player makes a given claim when they hold the role to how often they make
it when they do not. Calibrating against an honest claim rate near $0.85$ and a
bluff rate near $0.15$ gives a ratio of about $5.7$, which is applied when a role
is claimed through an action or through a Foreign Aid or Assassination block.
Steal blocks use $4.5$, because two different roles can block a Steal and the
claim therefore discriminates less. Surviving a challenge multiplies the claimed
role by $10.0$, since the card was shown, and losing one multiplies it by $0.05$.
Scores are additionally scaled by a \emph{deck pressure} factor $(c_r/3)^{0.7}$,
where $c_r$ is the number of copies of role $r$ not yet revealed dead, so that a
role whose copies are exhausted becomes unlikely regardless of what is claimed.
The exact constants and update code are given in Appendix~\ref{app:belief}.

\subsection{From Per-Card Belief to Per-Hand Probability}
A belief is a \emph{per-card} quantity, but a claim is a statement about a
\emph{hand}. If the cards were drawn independently, a player with $k_i$ remaining
influence would hold at least one copy of role $r$ with probability
\begin{equation}
p_{\mathrm{ind}}(i,r) = 1 - \bigl(1 - b_i(r)\bigr)^{k_i}.
\label{eq:ptruth}
\end{equation}
Cards are in fact dealt without replacement from a finite deck, so CoupVisor also
computes the \emph{hypergeometric} probability of holding at least one copy when
$k_i$ cards are drawn from a $15$-card pool containing $c_r$ copies:
\begin{equation}
p_{\mathrm{hyp}}(i,r) = 1 - \frac{\binom{15-c_r}{k_i}}{\binom{15}{k_i}}.
\label{eq:hyper}
\end{equation}
The deployed estimate blends the two, weighting the combinatorial term more
heavily and retaining the belief-derived term so that accumulated claim evidence
still moves the estimate in low-information games:
\begin{equation}
p_{\mathrm{truth}} = 0.7\,p_{\mathrm{hyp}} + 0.3\,p_{\mathrm{ind}}.
\label{eq:blend}
\end{equation}

\subsection{Threshold Rules and Expected Value}
A \emph{threshold rule} converts a probability into a binary decision by
comparing it against a cut-off: challenge when $p_{\mathrm{truth}} < \theta$,
otherwise do not. The threshold $\theta$ encodes how costly a wrong challenge is
relative to a missed one. The same decision can be written as an \emph{expected
value} (EV) comparison. Writing $G$ for the gain when a false claim is correctly
challenged and $L$ for the loss when a true claim is wrongly challenged, a
challenge is preferred when
\begin{equation}
(1 - p_{\mathrm{truth}})\,G \;>\; p_{\mathrm{truth}}\,L,
\label{eq:ev}
\end{equation}
which is a threshold rule with $\theta = G/(G+L)$. The threshold form is used in
the deployed advisor because it is inspectable, while the EV form is the
justification for the particular values chosen.

\subsection{Reinforcement Learning Terms}
\emph{Reinforcement learning} (RL) learns a \emph{policy} $\pi(a\mid s)$, a rule
mapping states to actions, by maximising the expected \emph{return}, the
discounted sum of future \emph{rewards} $\sum_t \gamma^t r_t$, where the
\emph{discount factor} $\gamma \in [0,1)$ trades immediate against future reward.
The \emph{action-value function} $Q(s,a)$ is the expected return from taking
action $a$ in state $s$ and behaving optimally thereafter, and \emph{Q-learning}
estimates it from observed transitions. A \emph{Deep Q-Network} (DQN) represents
$Q$ with a neural network and stabilises training with an \emph{experience replay
buffer}, which stores past transitions and samples them in random batches, and a
periodically synchronised \emph{target network}.

Two contrasts matter here. \emph{Behavior cloning} (BC) is supervised imitation:
it learns to predict the action a logged demonstrator took, with no notion of
reward. \emph{Offline RL} is reinforcement learning from a fixed, previously
collected set of transitions with no further interaction with the environment,
and its characteristic failure is \emph{distribution shift}, where the learned
policy proposes actions the logged data never covered. All learning in this paper
is offline, since policies are trained on simulator logs.

Finally, \emph{reward shaping} is the practice of adding intermediate reward
terms, here coin gain and influence retained, to a sparse objective in order to
give a denser learning signal. We contrast this with a \emph{win-aligned} reward
that credits the policy only when its chosen action matches the action taken by
the eventual winner of the logged game. Both reward functions appear verbatim in
Appendix~\ref{app:reward}.

\subsection{Exploration Versus Exploitation}
An \emph{exploitative} rule always takes the action that looks best under current
estimates, whereas an \emph{exploratory} rule sometimes takes an action whose
value is uncertain in order to improve those estimates and to avoid being
predictable. In an adversarial game the second concern is decisive: a
deterministic policy is \emph{exploitable}, because an opponent can infer it and
respond, so strong play requires \emph{mixed strategies} that randomise between
truthful actions, bluffs, and challenges. Three mechanisms appear in this paper.
\emph{$\varepsilon$-greedy} takes a uniformly random action with probability
$\varepsilon$ and the highest-valued action otherwise, annealing $\varepsilon$
over training. The \emph{Boltzmann} (softmax) policy
$\pi(a\mid s) = \mathrm{softmax}\bigl(Q(s,a)/\tau\bigr)$ spreads probability in
proportion to estimated value, with a \emph{temperature} $\tau$ controlling how
sharp the distribution is. \emph{Thompson sampling} draws a hypothesis from a
posterior distribution and acts as though it were true, so exploration is
allocated in proportion to genuine uncertainty rather than by a fixed schedule.
\emph{Entropy}, measured in nats, quantifies how spread out the resulting action
distribution is and is used below as a direct measurement of exploration.

\subsection{Evaluation Terms}
For the challenge decision, which is binary, \emph{precision} is the fraction of
recommended challenges that were correct, \emph{recall} is the fraction of false
claims that were challenged, and \emph{F1} is their harmonic mean.
\emph{Outcome accuracy} is the fraction of decision points at which the
recommendation matched the better outcome in hindsight. \emph{Calibration} asks
whether a stated probability matches observed frequency, so that a claim assigned
$0.7$ should be true about seventy percent of the time. It is the property most
directly implicated by our error analysis. A \emph{confidence interval} (CI)
gives the range in which a mean plausibly lies, and a difference whose $95\%$ CI
crosses zero is not claimed as reliable.

\section{Related Work}
The techniques used here rest on existing work in hidden-information decision
making, imitation learning, and reinforcement learning. Maintaining role
probabilities as a belief state follows the partial-observability framing of the
POMDP literature \cite{kaelbling1998pomdp}. The transition dataset is built as
Markov-style state, action, next-state tuples, and the RL objective follows
action-value learning \cite{watkins1992qlearning}. Policy learning from logged
behavior follows imitation-learning formulations \cite{ross2011dagger,
osa2018imitation}, with a Random Forest classifier \cite{breiman2001rf} used for
the supervised variant. The value-based baseline is DQN \cite{mnih2015dqn} with
experience replay \cite{lin1992replay}, and because training uses fixed logged
transitions without online interaction, the setting is offline RL
\cite{levine2020offline}. Evaluation uses precision, recall, F1, and
confusion-matrix statistics \cite{sokolova2009metrics}, ROC-AUC for binary tasks
\cite{fawcett2006roc}, and calibration-bin analysis for predicted probabilities
\cite{niculescu2005calibration}.

\section{Framework and Methodology}

\subsection{Overall Framework}
\begin{figure*}[!t]
\centering
\includegraphics[width=0.9\textwidth]{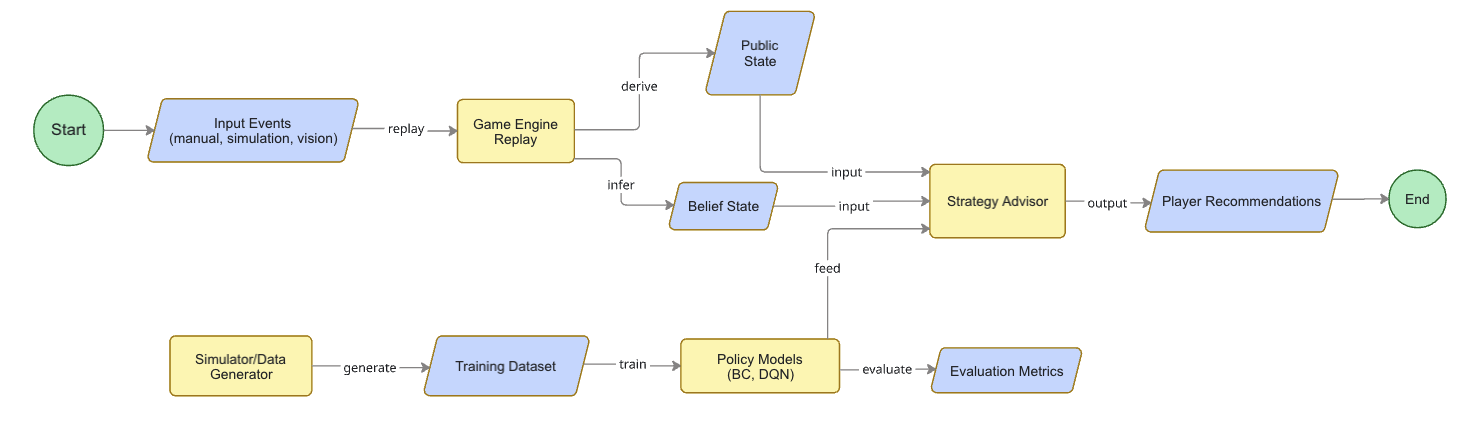}
\caption{The CoupVisor pipeline. Public events enter from any of three sources,
manual turn entry, replay of a recorded game, or the synthetic simulator, and are
applied by the engine, which maintains the public state: coins, remaining
influence, revealed roles, the current claim, and the last public action. The
belief tracker then updates the hidden-role distribution for every player, and
the advisor combines public state and belief into a player-facing recommendation.
Because all three input sources produce the same event schema, the advice path is
identical in live play, in replay, and during policy training.}
\label{fig:framework-diagram}
\end{figure*}

CoupVisor is organised as a single event pipeline, shown in
Fig.~\ref{fig:framework-diagram}. The same pipeline serves manual entry,
recorded-game replay, and the simulator, which is what allows training data and
live advice to remain consistent.

\subsection{Round Strategy Optimization Loop}
\begin{figure*}[!t]
\centering
\includegraphics[width=0.9\textwidth]{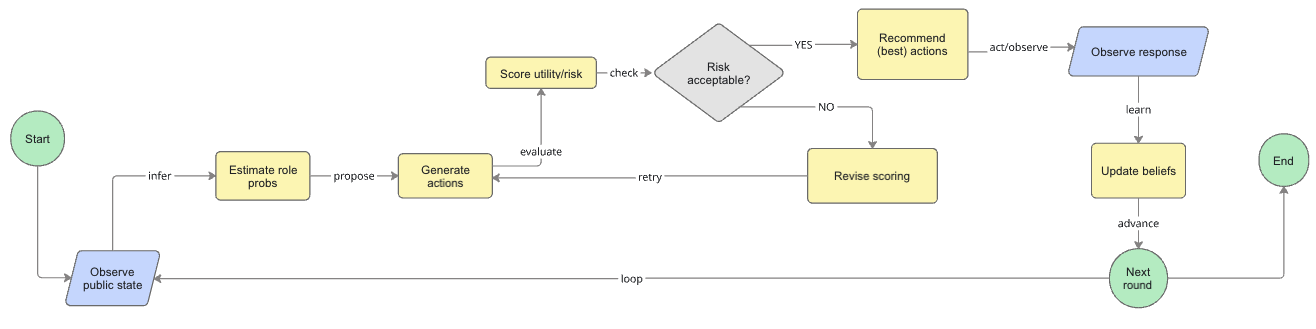}
\caption{Round-by-round strategy optimization loop. Each round begins by
estimating hidden roles from the current belief state, then enumerates the
actions that are legal given the player's coins and influence, scores each
candidate by expected utility against the risk that it is blocked or challenged,
and returns a ranked suggestion. The legality filter runs before scoring, so
constraints such as the forced Coup at ten coins remove options rather than
merely penalising them.}
\label{fig:strategy-loop}
\end{figure*}

The per-round loop, shown in Fig.~\ref{fig:strategy-loop}, is what produces
action advice as opposed to challenge advice. It is re-entered from scratch every
round rather than carrying a plan forward, because a single challenge can
invalidate any multi-turn intention.

\subsection{Challenge Decision Logic}
\begin{figure*}[!t]
\centering
\includegraphics[width=0.62\textwidth,height=0.82\textheight,keepaspectratio]{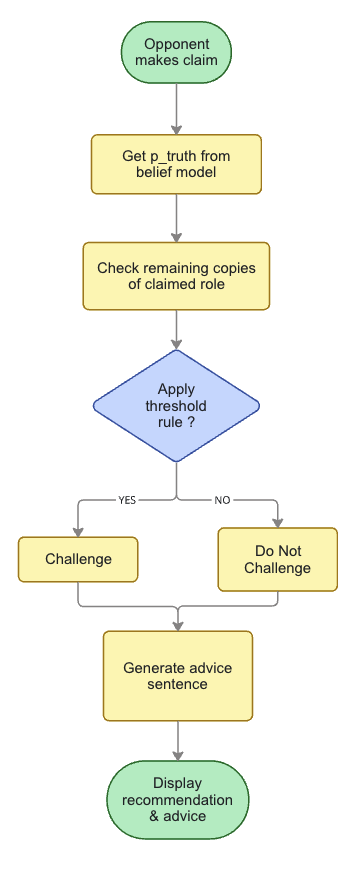}
\caption{Challenge decision logic. On observing a claim the advisor reads
$p_{\mathrm{truth}}$ from the belief model, checks how many copies of the claimed
role remain unaccounted for, and applies the threshold rule. The two branches
lead to \emph{Challenge} and \emph{Do not challenge}, each of which is turned
into a human-readable advice sentence carrying the probability and the threshold
that produced it, so the recommendation can be audited rather than merely
obeyed.}
\label{fig:challenge-logic}
\end{figure*}

When a player makes a role claim, the advisor computes $p_{\mathrm{truth}}$ from
(\ref{eq:blend}), compares it against a context-dependent threshold, and outputs
either \emph{Challenge} or \emph{Do not challenge}
(Fig.~\ref{fig:challenge-logic}).

Thresholds differ by claim context because the cost of being wrong differs. A
Steal block uses a base of $0.20$, an Assassinate action or an assassination block
uses $0.35$, a claim whose role has no copies left in the deck uses $0.99$, and
all other claims use $0.25$. The base is then adjusted by the user's risk mode and
by the state of the table. Fewer surviving players raises the threshold, by
$0.10$ at two players and $0.05$ at three, since with fewer opponents an
unchallenged bluff is more damaging. Holding a single influence lowers it by
$0.08$, since the challenger cannot afford to be wrong. Any opponent sitting at
seven or more coins raises it by $0.06$, since that opponent can Coup next turn
and the challenger is under time pressure. The result is clamped to
$[0.01, 0.99]$. The complete threshold function is reproduced in
Appendix~\ref{app:belief}.

The per-hand conversion is necessary rather than cosmetic. Without it, the
uniform turn-one per-card prior of $0.20$ sits below the Tax threshold of $0.25$
and produces a spurious \emph{Challenge} recommendation before any evidence
exists. Under (\ref{eq:ptruth}) alone the value would be
$1-(1-0.20)^2 \approx 0.36$, and under the deployed blend (\ref{eq:blend}) it is
higher still, so the opening claim correctly clears the threshold.

\subsection{Belief Update Logic}
\begin{figure*}[!t]
\centering
\includegraphics[width=0.55\textwidth,height=0.82\textheight,keepaspectratio]{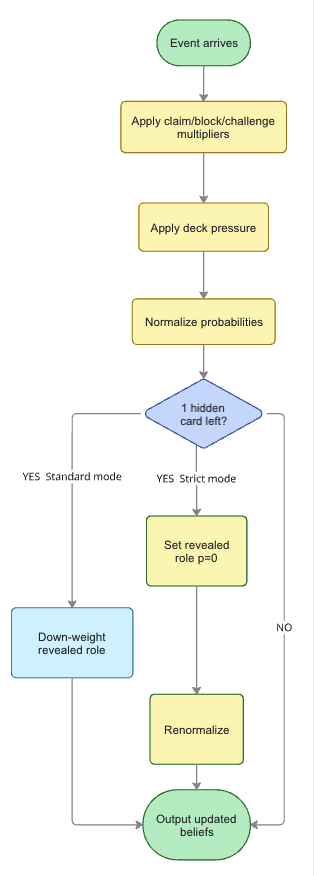}
\caption{Hidden-role belief update pipeline. An incoming event multiplies the
score of the role it implicates by the corresponding likelihood ratio, all scores
are then damped by deck pressure and normalised to a distribution, and finally
reveal constraints are applied. The two modes diverge only at the last step: in
\emph{Standard} mode duplicate roles in one hand are permitted, so a revealed
role can still carry non-zero hidden probability, whereas in \emph{Strict
no-duplicate-hand} mode a player down to a single hidden influence has the
revealed role forced to zero and the remainder renormalised.}
\label{fig:belief-diagram}
\end{figure*}

Belief updates apply the likelihood ratios and deck-pressure scaling defined in
Background, then normalise, then apply reveal constraints
(Fig.~\ref{fig:belief-diagram}). The two supported reveal modes differ in how
strictly they treat a revealed card, and the ablation in
Section~\ref{sec:ablation-belief} measures whether that difference matters.

\subsection{Coin Pressure and Legal Action Constraints}
Coins are not only a score variable, since they determine which actions are legal
and which threats are real. Each player starts with two coins, Assassinate
requires at least three, Coup requires at least seven, and a player holding ten
or more coins is compelled to Coup. The forced-Coup rule matters for both realism
and advisor behavior: once a player crosses ten coins the action space collapses
to a single legal action, so the uncertainty about their next move disappears and
a guaranteed attack enters the game. These constraints are enforced identically
in the simulator, in event validation, and in the interface event builder, which
keeps generated data, replayed data, and manually labelled data consistent.

\subsection{Advisor Reasoning Under Bluff Pressure}
Because Coup is a bluffing game, the advisor combines probabilistic and
rule-based evidence. The probabilistic part is the hidden-role estimate
$p_{\mathrm{truth}}$ for the claiming player, adjusted by the claim context and
by the number of copies of the claimed role still unaccounted for. The rule-based
part is a set of hard legality checks derived from coins and influence. Both are
shown to the user together, so the interface separates what is uncertain from
what is certain: belief expresses how likely the claim is, while regulation
expresses what is outright impossible. This separation is what makes the
recommendation auditable in a high-pressure decision.

The advisor exposes three user-facing risk modes. Conservative shifts the
threshold down by $0.05$ and preserves influence by challenging less often,
Aggressive shifts it up by $0.07$ and punishes likely bluffs more readily, and
Balanced is the unshifted default. All three share the same belief model and the
same legality constraints. In symmetric early-game states the belief values for
several players are close to uniform, which reflects genuinely insufficient
information rather than model failure. In those states the system reports the
uncertainty explicitly and offers conditional guidance, namely top warnings,
ranked candidate actions, and if-then paths, rather than a single deterministic
instruction.

Action advice is scored by a two-ply expected-value tree. For each legal action
the tree evaluates the immediate coin and influence swing, then the opponent
responses that the current belief makes likely, such as a Duke block against
Foreign Aid or a Contessa block against Assassinate, and finally the challenge
that could follow that response. Two plies are enough to capture the
block-and-challenge structure that decides most Coup turns without committing to
a full game-tree search over hidden states.

\subsection{Risk-Asymmetric Behavior in the Endgame}
Decision behavior becomes risk-asymmetric when a player has only one influence
left, because the cost of failure is already near its maximum. Such players bluff
and challenge more often than expected value alone would justify. For example, a
player facing an Assassinate with one influence and no Contessa may still claim
Contessa to block. The bluff is unlikely to survive a challenge, but the
alternative is certain elimination, so a low success probability is preferred to
none. Modelling this effect improves the realism of simulated opponents and
better matches observed human play.

A related pattern is deliberate loss for informational gain. Suppose Player~A
holds Assassin and believes Player~B holds Contessa. A may assassinate and then
challenge B's Contessa block. If A loses the challenge, A loses one influence but
survives, while B must reveal and redraw, and after the redraw B may no longer
hold Contessa. A has accepted a short-term loss to alter the opponent's role
distribution, raising the success probability of the next assassination. This is
belief manipulation rather than immediate value maximisation, and it is not
captured by a single-step decision rule.

\subsection{Thompson Sampling Extension}
Thompson sampling offers a principled alternative to both $\varepsilon$-greedy
exploration and fixed thresholds, and it unifies exploration across the belief
and action layers by drawing from posteriors rather than following schedules. At
the belief layer, instead of acting on posterior means, the system would sample a
latent role assignment for each opponent from the current belief and evaluate
candidate actions under that sampled world, so that repeated sampling naturally
balances optimistic and conservative readings of the same uncertainty. At the
challenge layer, it would maintain context-specific Beta posteriors over claim
truthfulness, sample $\theta \sim \mathrm{Beta}(\alpha,\beta)$ at decision time,
challenge when $(1-\theta)G > \theta L$, and update $(\alpha,\beta)$ from the
reveal outcome, giving adaptive behavior without global thresholds. At the action
layer, it would keep uncertain value estimates per legal action, sample one value
per action per turn, and select the highest sampled value subject to hard rules
such as forced Coup.

The three exploration mechanisms differ in how exploration is allocated.
$\varepsilon$-greedy explores uniformly at random, the Boltzmann policy explores
in proportion to the spread of estimated action values, and Thompson sampling
explores in proportion to posterior uncertainty over latent quantities. The last
is the most informed, at the cost of maintaining explicit posteriors.

\subsection{Decision Layers and Their Current Character}
CoupVisor makes decisions at three layers. The belief layer produces a single
normalised posterior per player and performs no sampling, in the spirit of
maximum a posteriori reasoning. The advisor layer applies thresholds and is
therefore exploitative: it maximises immediate expected value under the current
belief and returns the same recommendation whenever the belief and the table
state are the same. Only the learning layer explores, and it does so through
either $\varepsilon$-greedy or Boltzmann action selection during offline DQN
training.

\subsection{Relation to Poker-Style Imperfect Information}
Coup and poker share hidden information, bluff risk, and the need for mixed
strategies, and in both a strong agent must combine probability estimates with
action constraints. The difference is that Coup makes role claims and challenges
explicit and frequent, so challenge recommendation is a first-class decision in
essentially every round rather than an implicit consequence of betting. Coup's
coin constraints, particularly forced Coup at ten coins, also create hard legal
transitions stronger than typical poker betting constraints.

\subsection{Single-Turn Interaction}
\begin{figure*}[!t]
\centering
\includegraphics[width=0.9\textwidth]{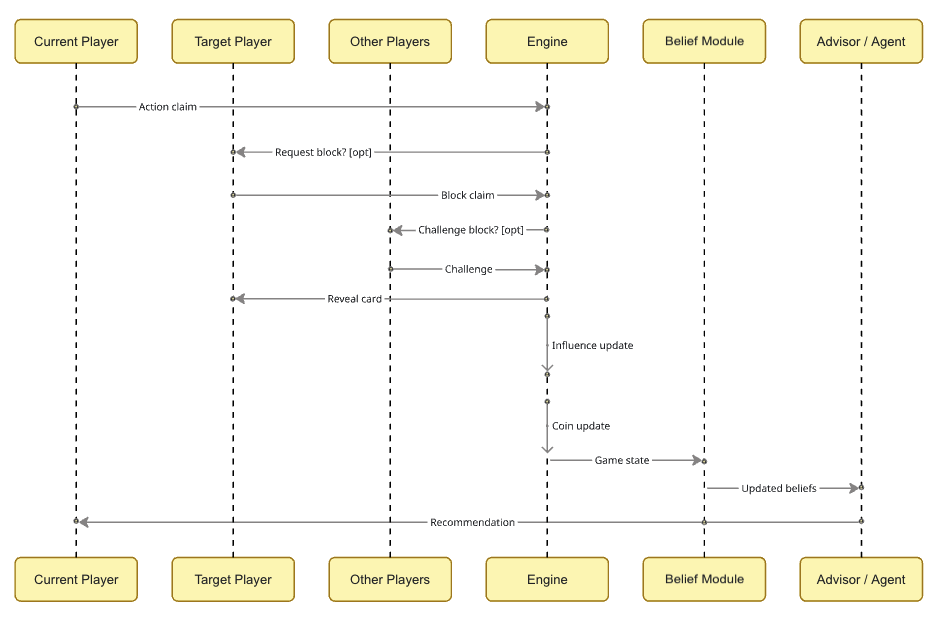}
\caption{Single-turn sequence. The turn opens with an action and its implied role
claim, after which the table may block, challenge the action, or challenge the
block. Each branch produces its own events, and a challenge resolves before the
action or counteraction it targets. Every arrow that carries a claim or a reveal
triggers a belief update and a fresh advisor recommendation, which is why a
single turn can generate several decision points rather than one.}
\label{fig:turn-sequence}
\end{figure*}

Fig.~\ref{fig:turn-sequence} shows how a single turn expands into the branches
that the event schema has to represent, and it is the unit that both the replay
engine and the simulator operate on.

\subsection{Worked Example}
A concrete four-player example illustrates how information accumulates. At the
start, Players A through D each hold two influence and two coins, nothing has
been revealed, and the belief state is uniform at $0.20$ per role
(Table~\ref{tab:belief-evolution}, Step~0).

Player A then claims Duke and takes Tax, moving to five coins. The claim raises
$P_A(\mathrm{Duke})$ and slightly depresses Duke for everyone else, since one of
the three copies is now plausibly accounted for (Step~1). At this point the
advisor evaluates the claim. It observes that no Duke has been revealed, that A
has shown no prior inconsistency, and that A holds two influence. It computes
$p_{\mathrm{truth}}$ from (\ref{eq:blend}) and compares expected outcomes through
(\ref{eq:ev}), weighing the card an opponent loses if the challenge succeeds
against the card the challenger loses if it fails. Suppose no one challenges. A
gains three coins, which reinforces but does not confirm the Duke hypothesis.

Later, A claims Assassin and targets B. Belief shifts toward Assassin for A
(Step~2), but the two claims are only jointly satisfiable if A holds exactly Duke
and Assassin, so the posterior mass on each individual claim being truthful falls
relative to a single-claim history. The advisor recomputes $p_{\mathrm{truth}}$
under the new deck composition and the accumulated claim history, and issues a
recommendation for whether B should challenge or block.
Table~\ref{tab:example-summary} summarises the sequence.

\begin{table*}[!t]
\centering
\caption{Belief evolution over the worked example. Step~0 is the uniform prior,
Step~1 follows A's Duke claim, Step~2 follows A's Assassin claim.}
\label{tab:belief-evolution}
\begin{tabular}{llccccc}
\toprule
Step & Player & $P(\mathrm{Duke})$ & $P(\mathrm{Assassin})$ & $P(\mathrm{Captain})$ & $P(\mathrm{Ambassador})$ & $P(\mathrm{Contessa})$ \\
\midrule
0 & A & 0.20 & 0.20 & 0.20 & 0.20 & 0.20 \\
0 & B, C, D & 0.20 & 0.20 & 0.20 & 0.20 & 0.20 \\
\midrule
1 & A & 0.55 & 0.15 & 0.10 & 0.10 & 0.10 \\
1 & B, C, D & 0.15 & 0.20 & 0.20 & 0.20 & 0.25 \\
\midrule
2 & A & 0.30 & 0.40 & 0.10 & 0.10 & 0.10 \\
2 & B & 0.15 & 0.15 & 0.20 & 0.20 & 0.30 \\
2 & C, D & 0.15 & 0.20 & 0.20 & 0.20 & 0.25 \\
\bottomrule
\end{tabular}
\end{table*}

\begin{table}[!t]
\centering
\caption{Event sequence and corresponding advisor update.}
\label{tab:example-summary}
\begin{tabular}{@{}cp{0.33\columnwidth}p{0.36\columnwidth}@{}}
\toprule
Step & Event & Advisor update \\
\midrule
1 & A claims Duke (Tax) & Increase $P_A(\mathrm{Duke})$ \\
2 & No challenge & Belief remains uncertain \\
3 & A claims Assassin & Reduce joint-claim consistency \\
4 & Decision point & Compute $p_{\mathrm{truth}}$, compare EV \\
\bottomrule
\end{tabular}
\end{table}

\subsection{Training and Evaluation Pipeline}
\begin{figure*}[!t]
\centering
\includegraphics[width=0.9\textwidth]{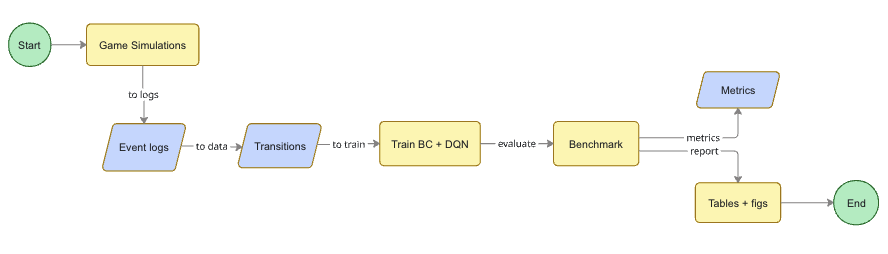}
\caption{Training and evaluation pipeline. Simulated games are flattened into
state, action, reward, next-state transitions, which feed both the behavior
cloning head and the offline DQN. Both learned policies and the non-learning
baselines are then replayed through the same evaluation harness on the same
transitions, so that reward per episode, reward per step, and action-match rate
are computed identically for every policy.}
\label{fig:train-eval-pipeline}
\end{figure*}

Simulator output is converted into transition datasets used to train the BC and
DQN policies, which are then benchmarked against a random policy and against the
non-learning agents described below (Fig.~\ref{fig:train-eval-pipeline}).

\subsection{Non-Learning Baseline Agents}
Random action selection is too weak to be informative on its own, so three
non-learning agents of increasing decision quality are used as reference points.
The \emph{rule-based legal-priority agent} enforces legality first, including
forced Coup at ten coins, then follows a fixed preference order of Coup,
Assassinate, Steal, Tax, Foreign Aid, and finally Income or Exchange. Its
challenge decisions use hand-crafted thresholds keyed to claim context and
remaining deck copies. The \emph{belief-based expected-value agent} shares the
advisor's belief estimator but has no trainable parameters. For each legal action
it computes a one-step expected value from coin swing, influence swing, and the
block or challenge response probabilities implied by current beliefs, and it
challenges according to (\ref{eq:ev}). The \emph{public-state pressure agent}
ignores hidden-role beliefs entirely and uses only public features, namely coin
pressure, influence counts, and an opponent threat ranking. It targets high-coin
opponents and challenges mainly in high-impact contexts such as lethal
Assassinate blocks or exhausted role copies. Together these span belief-free,
belief-driven, and purely procedural decision making.

\subsection{User Interaction Flow}
\begin{figure*}[!t]
\centering
\includegraphics[width=0.9\textwidth]{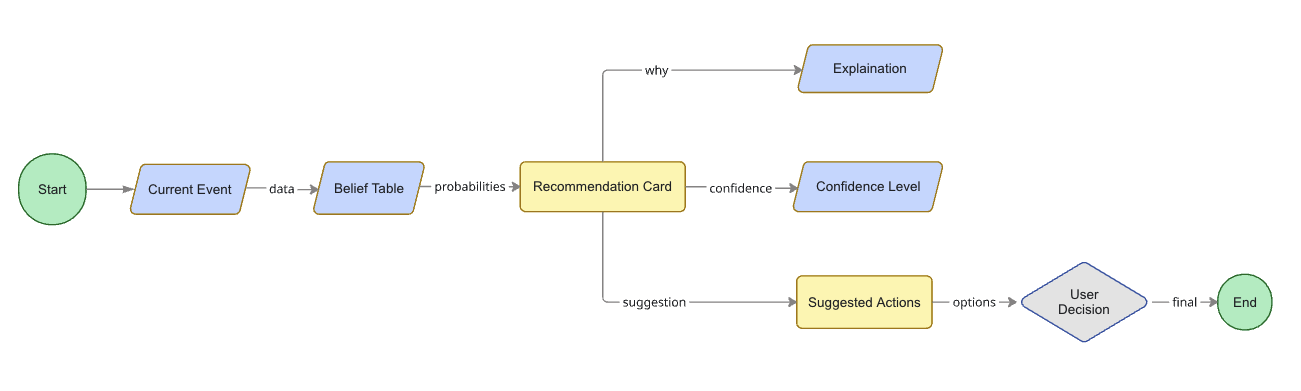}
\caption{User interaction flow. The user loads or enters a game, steps through it
event by event, and at each step sees the reconstructed public state, the
hidden-role belief table, and the advisor recommendation for any claim currently
on the table. Advice and validation warnings are presented side by side, so that
a probabilistic judgement is never confused with a hard rule violation.}
\label{fig:ui-flow}
\end{figure*}

Fig.~\ref{fig:ui-flow} shows the user-facing path, which is deliberately shallow
so that a non-technical user never has to interact with the event schema
directly.

\subsection{Reproduction}
The components that produce every number reported here are given verbatim in the
appendices: the belief update and threshold rule in Appendix~\ref{app:belief},
the two reward definitions in Appendix~\ref{app:reward}, and the network,
optimiser, and training loop in Appendix~\ref{app:train}.

\section{Experimental Setup}

\subsection{Evaluation Run}
All numbers and plots in this paper come from a single evaluation run whose
configuration, seeds, and outputs were fixed before analysis, so that every
table, ablation, and figure describes the same set of games.

\subsection{Primary Design Matrix}
The main experiment uses a balanced matrix so that conditions are directly
comparable. Each of five seeds contributes the same number of games to each
opponent condition (Table~\ref{tab:exp-matrix-primary}), for 250 games per
condition and 500 games in total.

\begin{table}[!t]
\centering
\caption{Primary balanced matrix (seed $\times$ condition).}
\label{tab:exp-matrix-primary}
\begin{tabular}{lcc}
\toprule
Seed & Aggressive-heavy & Honest-heavy \\
\midrule
101 & 50 games & 50 games \\
202 & 50 games & 50 games \\
303 & 50 games & 50 games \\
404 & 50 games & 50 games \\
505 & 50 games & 50 games \\
\midrule
Total & 250 games & 250 games \\
\bottomrule
\end{tabular}
\end{table}

\subsection{Metrics}
Advisor quality is measured by challenge precision, challenge recall, challenge
F1, outcome accuracy, and outcome score. Policy quality is measured by average
reward per episode, average reward per step, and action-match rate against the
logged reference action.

\subsection{Exploration-Mode Ablation}
For the DQN training protocol we additionally ablate the action-selection rule.
Two conditions are compared under identical seeds, network architecture, replay
buffer, and reward function: $\varepsilon$-greedy with
$\varepsilon_{\mathrm{start}}=1.0$, $\varepsilon_{\mathrm{end}}=0.05$, and
multiplicative decay $0.995$ per episode, and a Boltzmann softmax policy with
$\tau_{\mathrm{start}}=1.0$, $\tau_{\mathrm{end}}=0.1$, and the same decay. For
each condition we record the running average episode reward, the mean per-step
entropy of the action distribution in nats, and the action-visitation histogram.
The entropy curve measures the realised exploration schedule, which a scalar
$\varepsilon$ or $\tau$ value alone does not capture.

\subsection{Evaluation Beyond Random Comparison}
Random-policy comparison is retained as a weak lower bound, but the primary
evaluation uses three stronger protocols. \emph{Scenario-based evaluation} runs
all policies on fixed decision scenarios that isolate the core Coup conflicts,
namely early Tax claims, Steal-block-challenge branches, Assassinate-Contessa
branches, forced-Coup states, and one-influence endgames, under shared seeds,
which removes rollout variance and permits decision-level comparison.
\emph{Opponent-type robustness} runs cross-play against honest-heavy, bluff-heavy,
aggressive-heavy, cautious-heavy, and mixed opponent pools, reported as a matrix
over train and test opponent types to expose behavioral shift.
\emph{Decision-quality metrics} supplement reward with challenge precision,
recall, and F1, calibration quality such as Brier score, legality rate,
forced-Coup compliance, and one-step regret against a belief-based oracle
approximation, so that strategic soundness is assessed separately from final
reward.

\section{Results}

\subsection{Advisor Multi-Seed Results}
\label{sec:advisor-multiseed}
Table~\ref{tab:advisor-main} reports advisor metrics aggregated over the ten
seed-condition cells. Challenge F1 has a wide interval, $0.336$ to $0.511$, while
outcome accuracy is comparatively stable at $0.5694 \pm 0.0242$. The contrast
matters: the outcome-level behavior of the advisor is stable across seeds, but
its challenge decisions are not, and the variance is driven by opponent
composition rather than by seed noise, as the next subsection shows.

\begin{table}[!t]
\centering
\caption{Advisor metrics across seed-condition cells.}
\label{tab:advisor-main}
\begin{tabular}{lcccc}
\toprule
Metric & Mean & Std & CI low & CI high \\
\midrule
Challenge F1 & 0.4236 & 0.1409 & 0.3363 & 0.5110 \\
Outcome accuracy & 0.5694 & 0.0242 & 0.5544 & 0.5844 \\
Outcome score & 0.1428 & 0.0486 & 0.1127 & 0.1730 \\
\bottomrule
\end{tabular}
\end{table}

\subsection{Bot-Mix Sensitivity}
Table~\ref{tab:botmix} separates the two opponent conditions. Challenge F1 is
almost twice as high against aggressive-heavy opponents ($0.5641$) as against
honest-heavy opponents ($0.2831$), which is the expected direction: when few
claims are false, recall has little to find and precision is punished by every
challenge. Outcome accuracy moves in the opposite direction but only slightly,
from $0.5619$ to $0.5769$. Table~\ref{tab:ranking} gives the weighted ranking
score, $0.5628$ against aggressive-heavy and $0.4594$ against honest-heavy.

\begin{table}[!t]
\centering
\caption{Condition comparison by bot mix (mean across seeds).}
\label{tab:botmix}
\begin{tabular}{lccc}
\toprule
Condition & Challenge F1 & Outcome acc. & Outcome score \\
\midrule
Aggressive-heavy & 0.5641 & 0.5619 & 0.1239 \\
Honest-heavy & 0.2831 & 0.5769 & 0.1617 \\
\bottomrule
\end{tabular}
\end{table}

\begin{table}[!t]
\centering
\caption{Weighted ranking score by condition.}
\label{tab:ranking}
\begin{tabular}{lc}
\toprule
Condition & Weighted score \\
\midrule
Aggressive-heavy & 0.5628 \\
Honest-heavy & 0.4594 \\
\bottomrule
\end{tabular}
\end{table}

\subsection{Ablation: Belief Mode}
\label{sec:ablation-belief}
Table~\ref{tab:ablation-belief} compares Standard against Strict
no-duplicate-hand belief updating. All three deltas are small and negative, and
all three $95\%$ confidence intervals cross zero, so no reliable difference is
claimed. The practical reading is that the duplicate-hand constraint rarely binds
at the decision points that matter, because by the time a player is down to one
hidden influence the belief is usually already dominated by other evidence.

\begin{table}[!t]
\centering
\caption{Standard vs strict no-duplicate-hand belief mode.}
\label{tab:ablation-belief}
\begin{tabular}{lccc}
\toprule
Mode & Challenge F1 & Outcome acc. & Outcome score \\
\midrule
Standard & 0.4236 & 0.5694 & 0.1428 \\
Strict no-duplicate & 0.4190 & 0.5684 & 0.1408 \\
$\Delta$ (Strict $-$ Std.) & $-0.0046$ & $-0.0010$ & $-0.0020$ \\
\bottomrule
\end{tabular}
\end{table}

\subsection{Ablation: Advisor On Versus Off}
Table~\ref{tab:ablation-advisor} compares the advisor against a never-challenge
baseline. The confidence intervals are far from zero on all three metrics, but
the direction requires care. Turning the advisor off drives challenge F1 to zero
by construction, yet raises outcome accuracy to $0.8192$ and outcome score to
$0.6424$. This is not evidence that challenging is harmful. It is evidence that
in this simulator most claims are true, so a policy that never challenges is
rarely punished. The comparison establishes the cost side of challenging, not its
overall value, and it should not be read as a full strategic baseline.

\begin{table}[!t]
\centering
\caption{Advisor policy ablation (off = never-challenge baseline).}
\label{tab:ablation-advisor}
\begin{tabular}{lccc}
\toprule
Mode & Challenge F1 & Outcome acc. & Outcome score \\
\midrule
Advisor on & 0.4236 & 0.5694 & 0.1428 \\
Advisor off & 0.0000 & 0.8192 & 0.6424 \\
$\Delta$ (Off $-$ On) & $-0.4236$ & $+0.2498$ & $+0.4996$ \\
\bottomrule
\end{tabular}
\end{table}

\subsection{Significance Notes}
Every Strict-minus-Standard interval crosses zero, so that difference is not
claimed. Every advisor-on-minus-off interval excludes zero, so those differences
are real in the statistical sense, subject to the interpretive caveat above.

\subsection{Policy Benchmark}
\label{sec:policy-benchmark}
Table~\ref{tab:policy} reports the benchmark under the original shaped reward,
which sums coin change, influence change, and an action-match bonus. Behavior
cloning reaches $+1.04$ average reward per episode against $-0.29$ for DQN and
$-0.60$ for random. The DQN reward curve in Fig.~\ref{fig:dqn-curve} oscillates
between roughly $-6$ and $+4$ per episode with a rolling mean band of $[-3,+0.5]$
and no upward trend across 100 episodes.

\begin{table}[!t]
\centering
\caption{Policy benchmark under the original shaped reward
($\Delta$coins $+$ $\Delta$alive $+$ action-match).}
\label{tab:policy}
\begin{tabular}{lccc}
\toprule
Policy & Reward/episode & Reward/step & Action match \\
\midrule
DQN & $-0.2868$ & $-0.0179$ & 0.1905 \\
Random & $-0.6004$ & $-0.0375$ & 0.1415 \\
Behavior cloning & $1.0412$ & $0.0651$ & 0.3980 \\
\bottomrule
\end{tabular}
\end{table}

Three factors explain the plateau. The training budget is short. The per-step
shaping mixes coin change, influence change, and a $+1$ action-match bonus, so
variance from logged-bot randomness dominates any long-horizon signal. The
offline environment replays transitions from heterogeneous bot styles rather than
from winning trajectories only. Since the downstream goal is to win the game
rather than to accumulate coins, we retrained DQN with a win-aligned reward that
grants $+1$ only when the chosen action matches the action taken by the eventual
winner, and extended training to 1000 episodes.

Table~\ref{tab:policy-v2} reports the revised benchmark, which also adds an
\emph{Honest} baseline that takes Tax when poor, Assassinate when able, and Coup
when rich. The ordering inverts completely. DQN now leads at $6.74$ reward per
episode and $0.373$ winner-action match, ahead of the belief-EV heuristic
($3.68$), behavior cloning ($3.38$), and the remaining heuristics, with random at
$1.32$. The corresponding reward curve, Fig.~\ref{fig:dqn-curve-v2}, rises from a
rolling mean near $1.5$ at episode~20 to roughly $5.9$ by episode~1000.
Table~\ref{tab:train-summary} shows the same trend in the training summary, where
average episode reward grows monotonically from $12.7$ at 10 episodes to $22.9$
at 5000 while episode length stays fixed at $24.46$ steps. Learned policies
therefore improve over random under both rewards, but which learned policy is
stronger is decided by the reward definition, not by the algorithm.

\begin{table}[!t]
\centering
\caption{Policy benchmark under win-aligned reward ($+1$ when the chosen action
matches the winner's logged action). 100 games; 1000 training episodes for DQN.}
\label{tab:policy-v2}
\begin{tabular}{lccc}
\toprule
Policy & Reward/episode & Reward/step & Winner match \\
\midrule
DQN (win-reward) & \textbf{6.74} & \textbf{0.248} & \textbf{0.373} \\
Belief-EV heuristic & 3.68 & 0.135 & 0.263 \\
Behavior cloning & 3.38 & 0.124 & 0.330 \\
Honest (baseline) & 2.84 & 0.104 & 0.244 \\
Rule-based heuristic & 2.42 & 0.089 & 0.264 \\
Pressure heuristic & 2.30 & 0.085 & 0.245 \\
Random & 1.32 & 0.049 & 0.141 \\
\bottomrule
\end{tabular}
\end{table}

\begin{table}[!t]
\centering
\caption{DQN training summary under the win-aligned reward.}
\label{tab:train-summary}
\begin{tabular}{lccc}
\toprule
Episodes & Avg ep.\ reward & Avg ep.\ steps & Final $\varepsilon$ \\
\midrule
10 & 12.700 & 22.60 & 0.95111 \\
50 & 14.580 & 24.46 & 0.77213 \\
100 & 14.840 & 24.46 & 0.60577 \\
200 & 15.045 & 24.46 & 0.36695 \\
500 & 16.278 & 24.46 & 0.08157 \\
1000 & 17.793 & 24.46 & 0.00500 \\
5000 & 22.901 & 24.46 & $\leq 0.005$ \\
\bottomrule
\end{tabular}
\end{table}

\subsection{Result Figures}
\begin{figure*}[!t]
\centering
\includegraphics[width=0.9\textwidth]{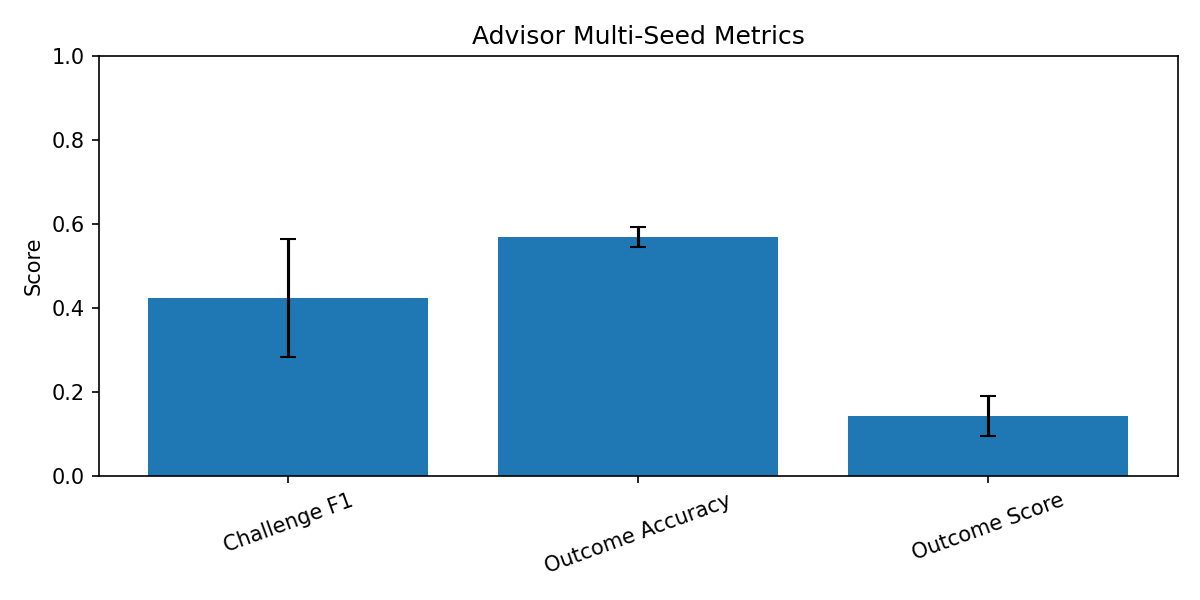}
\caption{Advisor metrics across the ten seed-condition cells. Each metric is
plotted with its spread over cells, which makes visible the contrast reported in
Table~\ref{tab:advisor-main}: outcome accuracy sits in a narrow band while
challenge F1 varies widely, so the instability is confined to the challenge
decision rather than affecting the advisor as a whole.}
\label{fig:advisor-artifact}
\end{figure*}

\begin{figure*}[!t]
\centering
\includegraphics[width=0.9\textwidth]{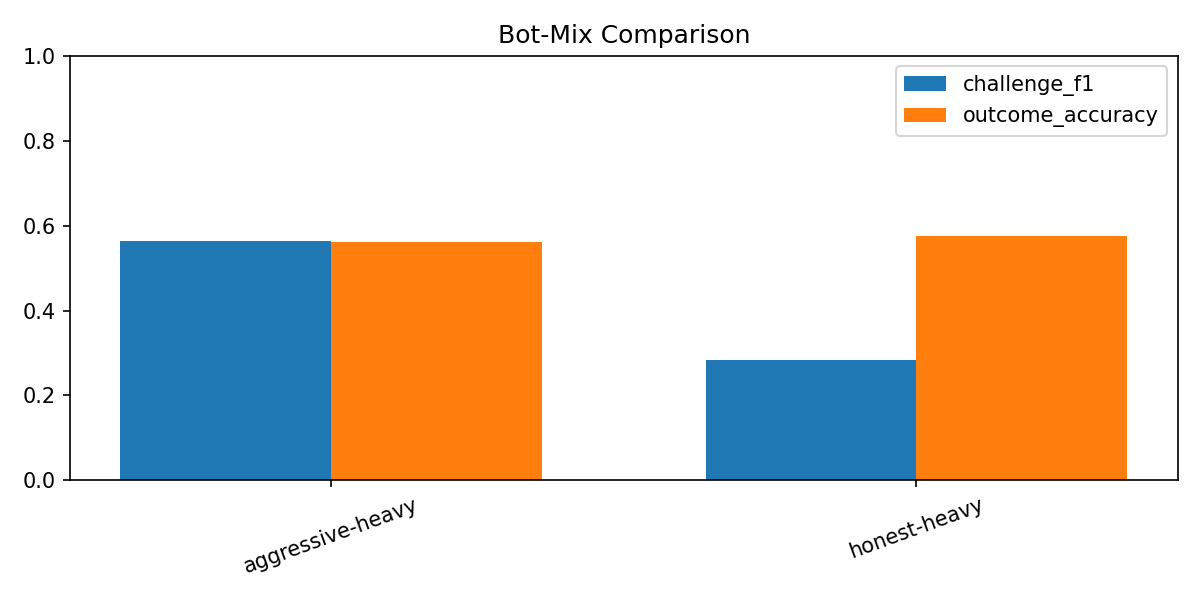}
\caption{Advisor performance split by opponent condition. The aggressive-heavy
and honest-heavy pools are shown side by side for each metric. Challenge F1
separates sharply between the two while outcome accuracy barely moves,
identifying opponent composition as the source of the F1 variance in
Fig.~\ref{fig:advisor-artifact}.}
\label{fig:botmix-artifact}
\end{figure*}

\begin{figure*}[!t]
\centering
\includegraphics[width=0.9\textwidth]{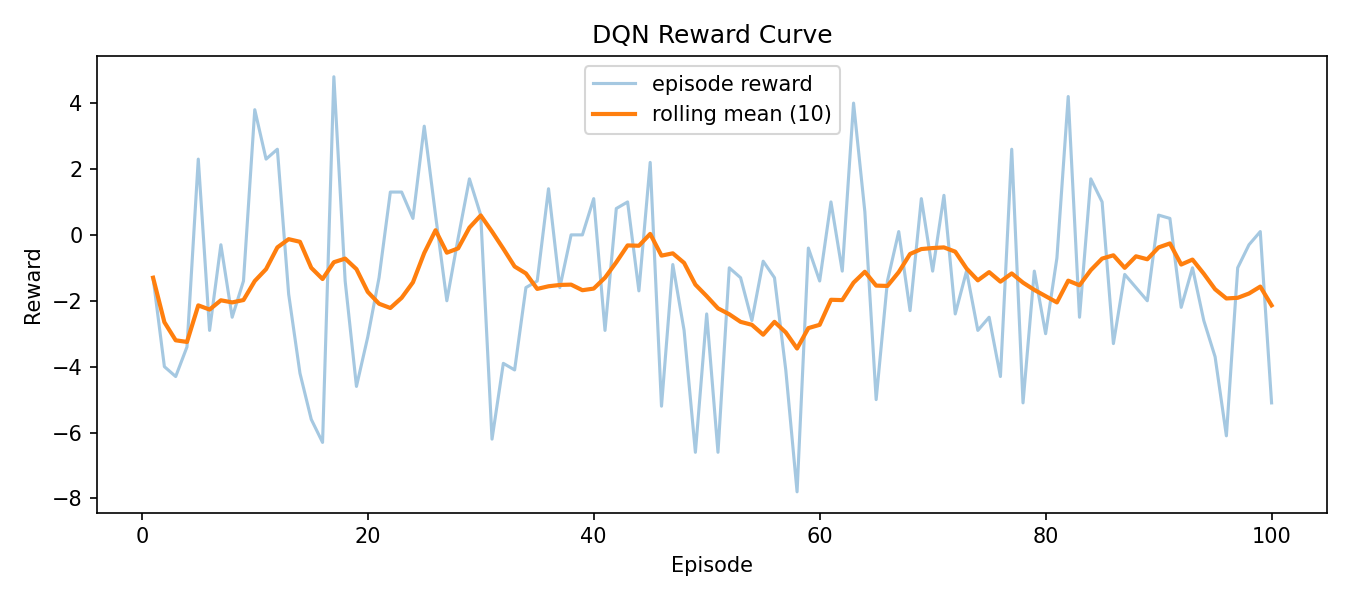}
\caption{DQN reward curve under the original shaped reward, 100 episodes.
Per-episode reward oscillates between roughly $-6$ and $+4$, and the rolling mean
stays inside $[-3,+0.5]$ with no upward trend, so the policy is not learning.
This flat curve is what motivated changing the reward rather than the algorithm.
Corresponds to Table~\ref{tab:policy}.}
\label{fig:dqn-curve}
\end{figure*}

\begin{figure*}[!t]
\centering
\includegraphics[width=0.9\textwidth]{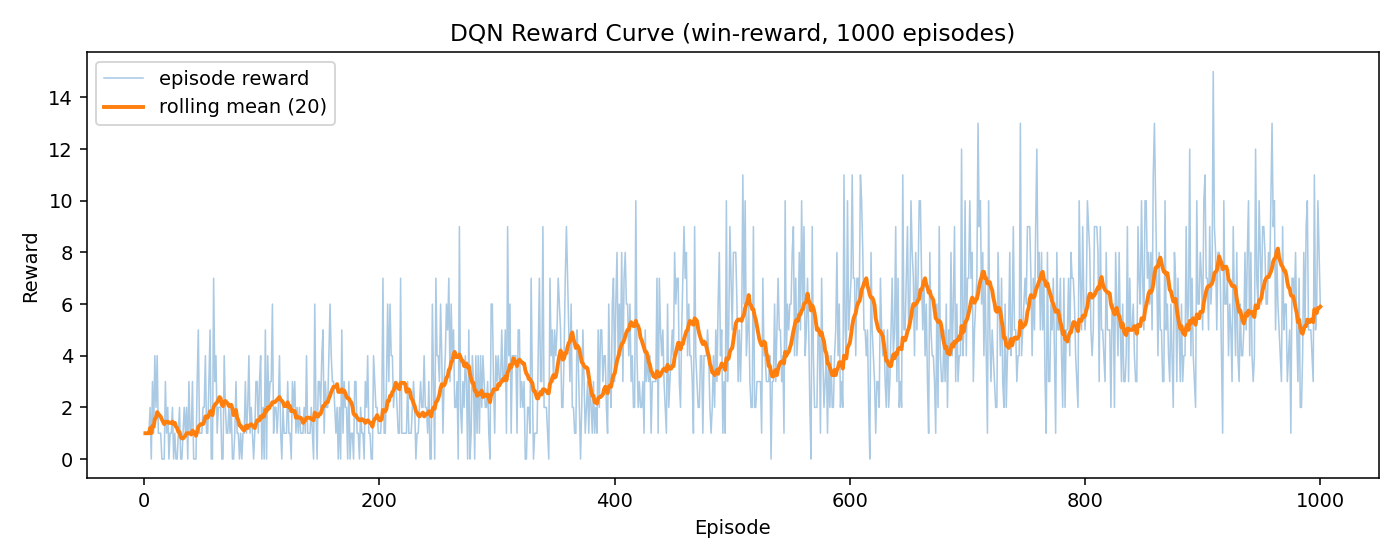}
\caption{DQN reward curve under the win-aligned reward, 1000 episodes. The
rolling mean rises from about $1.5$ at episode~20 to about $5.9$ by
episode~1000, a steady climb rather than the flat band of
Fig.~\ref{fig:dqn-curve}, confirming that the same architecture does learn once
the reward credits winner-matching actions. Corresponds to
Table~\ref{tab:policy-v2}.}
\label{fig:dqn-curve-v2}
\end{figure*}

\begin{figure*}[!t]
\centering
\includegraphics[width=0.9\textwidth]{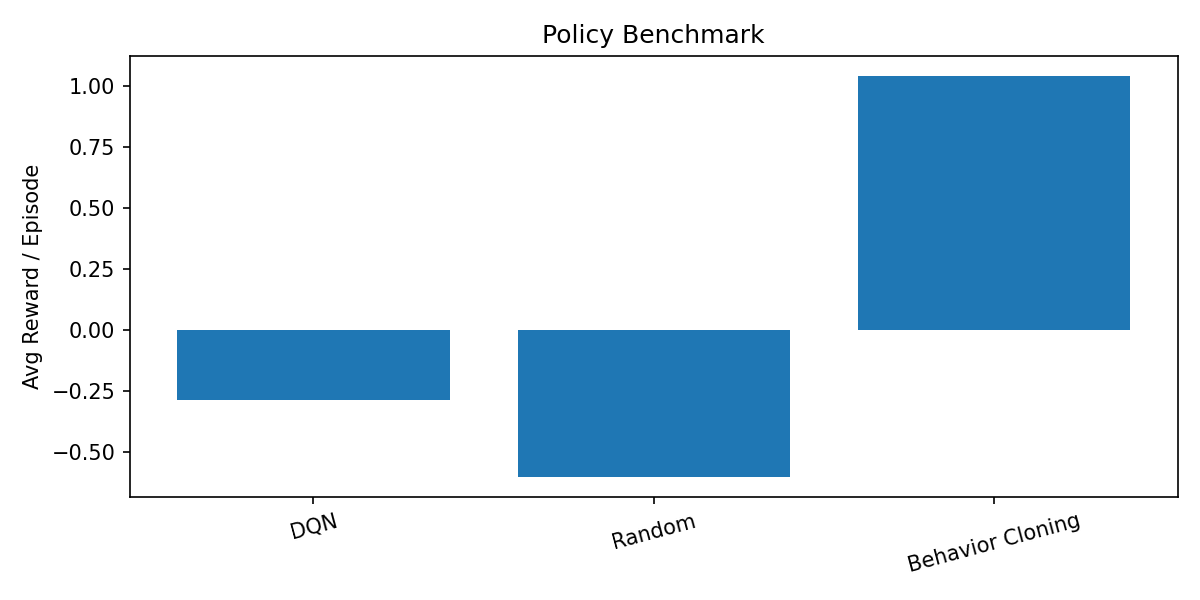}
\caption{Policy benchmark summary. Every policy is scored on the same
transitions, so the bars are directly comparable. Under the shaped reward the
learned and heuristic policies cluster near or below zero while behavior cloning
leads, the ordering that the win-aligned rerun in Table~\ref{tab:policy-v2}
subsequently inverts.}
\label{fig:policy-artifact}
\end{figure*}

Figures~\ref{fig:advisor-artifact} through~\ref{fig:policy-artifact} present the
same results graphically.

\subsection{Exploration-Mode Comparison}
\begin{figure*}[!t]
\centering
\includegraphics[width=0.9\textwidth]{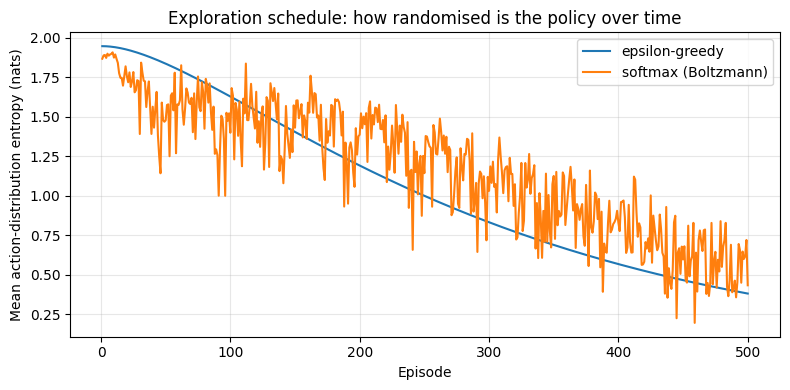}
\caption{Exploration-mode ablation under identical seeds, architecture, buffer,
and reward. Left: running average episode reward. Middle: mean per-step entropy
of the action distribution in nats. Right: action-visitation histogram. The
$\varepsilon$-greedy entropy curve is monotone in $\varepsilon$ and so is fixed
entirely by the schedule, whereas the Boltzmann entropy curve also responds to
the spread of learned action values and can plateau above the schedule floor when
several actions have comparable value. The two mechanisms therefore allocate
exploration differently even when their scalar schedules agree, which is why
entropy rather than the schedule parameter is the quantity worth reporting.}
\label{fig:exploration-modes}
\end{figure*}

Fig.~\ref{fig:exploration-modes} contrasts the two exploration modes and shows
that matching the scalar schedules does not match the exploration actually
performed.

\section{Discussion}

\subsection{What the Results Say}
The advisor produces usable recommendations but is imbalanced between precision
and recall, and the imbalance is systematic rather than random: it is a
calibration problem, not a modelling failure. Opponent composition changes
advisor performance substantially, which means any single reported F1 is only
meaningful alongside the opponent mix it was measured against. Learned policies
beat random under both reward definitions, but the ranking among them inverts
when the reward changes. Coin pressure creates a sharp structural break in the
game, since past ten coins the forced Coup collapses next-action uncertainty to
zero.

One design property deserves explicit statement. The deployed advisor is a
deterministic threshold rule and therefore sits at the exploitation end of the
spectrum. This was chosen for interpretability, and it is in tension with the
mixed-strategy requirement of imperfect-information games, which predicts that
any deterministic rule is exploitable in the limit. The exploration-mode ablation
and the Thompson-sampling extension both point to the same remedy, namely a
stochastic advisor whose mixing probabilities come from posterior uncertainty
rather than from hand-chosen constants.

\subsection{How This Research Helps}
The clearest practical contribution is to the player. A challenge decision is
normally made from intuition under time pressure, and the intuition is difficult
to inspect afterwards. CoupVisor replaces it with a stated probability, the
evidence that produced it, and a separate statement of what the rules make
impossible. That separation, uncertain belief on one side and hard legality on
the other, is what makes the advice teachable. A learner can see that a
recommendation not to challenge came from a per-hand probability rather than a
per-card one, and can see when the recommendation was driven purely by deck
exhaustion. The three risk modes let the same evidence be read under different
tolerances, which is closer to how humans actually differ than a single optimal
answer would be.

The second contribution is methodological, and it is the finding we consider most
transferable. Under a shaped short-term reward, behavior cloning beat DQN by a
wide margin. Under a reward aligned with winning, the same DQN architecture beat
every baseline including behavior cloning. Nothing about the network, the buffer,
or the data changed. The lesson generalises to any offline decision-support
system trained from logged behavior: when the logged population is heterogeneous
in quality, shaping toward local proxies teaches the model to imitate the average
participant, whereas conditioning the reward on the outcome you actually care
about teaches it to imitate the successful ones. Practitioners who observe a
flat, non-improving offline RL curve should audit the reward definition before
concluding that the algorithm or the data is inadequate.

The third contribution is architectural. Building every component on one event
schema, covering manual entry, replay, simulation, belief tracking, advice, and
training, means a change to the decision rule propagates to live advice, replay,
and training simultaneously, with no risk that the deployed rule and the
evaluated rule drift apart. This is the property that most decision-support
systems lose first, and the Coup setting is small enough to demonstrate it end to
end.

The fourth contribution is the structure of the problem itself. The challenge
decision, in which one pays a cost to verify a claim that may be false and is
punished for being wrong in either direction, has the same structure as an
inspection or audit decision. Deciding whether to audit a filing, verify a
self-reported metric, escalate a flagged transaction, or request proof in a
negotiation all share the form of (\ref{eq:ev}): a posterior over truthfulness,
an asymmetric cost, and a threshold that has to be set deliberately. Coup
provides a fully observable, cheaply simulated instance of that structure in
which belief updating and threshold setting can be varied and measured
independently, which is difficult in the real settings it resembles.

Finally, the error analysis below converts these results into specific
engineering actions rather than general advice, and because the evaluation set is
fixed, each of those actions can be assessed against exactly the same 500 games.

\subsection{Error Analysis}
We inspected all 2,670 challenged claims in the evaluation run. Two failure
groups dominate. In 942 cases ($35.3\%$) the advisor recommended \emph{Challenge}
but the claim was true, and in 96 cases ($3.6\%$) it recommended \emph{Do not
challenge} but the claim was false. The asymmetry is the numeric form of the
precision-recall imbalance noted above: the system challenges far too eagerly. By
action type, false challenges concentrate on Steal (467 of 942) and Tax (299 of
942), while missed challenges are mostly Steal (62 of 96).
Table~\ref{tab:error-cases} gives representative cases.

\begin{table*}[!t]
\centering
\caption{Representative failure cases from the evaluation run.}
\label{tab:error-cases}
\small
\begin{tabular}{@{}p{0.05\textwidth} p{0.20\textwidth} p{0.26\textwidth} p{0.40\textwidth}@{}}
\toprule
Case & Evidence & System output & Ground truth and issue \\
\midrule
E1 & game 339, event 48 (Tax, Duke by P2) & \emph{Challenge}; $p_{\mathrm{truth}}=1.93\mathrm{e}{-34}$, thr.\ $0.25$ & Challenge result is a win for the claimant, so the claim was true. Overconfident false positive. \\
E2 & game 231, event 1 (Steal block, Ambassador by P2) & \emph{Do not challenge}; $p_{\mathrm{truth}}=0.20$, thr.\ $0.20$ & Claimant loses, so the claim was false. Borderline equality missed a profitable challenge. \\
E3 & game 158, event 7 (Steal, Captain by P3) & \emph{Do not challenge}; $p_{\mathrm{truth}}=0.2503$, thr.\ $0.25$ & Claimant loses, so the claim was false. Near-threshold miss. \\
E4 & game 417, event 49 (Foreign Aid block, Duke by P4) & \emph{Challenge}; $p_{\mathrm{truth}}=8.98\mathrm{e}{-34}$, thr.\ $0.35$ & Claim was true. Block-claim handling is calibration-sensitive. \\
E5 & DQN trace, game 106, step 12 & DQN chose \emph{Exchange} holding 11 coins & Legal action is \emph{Coup}. In 10+ coin states DQN chose non-Coup 151 of 181 times, so forced-Coup alignment is weak. \\
\bottomrule
\end{tabular}
\end{table*}

Cases E2 and E3 sit within $0.0003$ of their threshold, so a small margin band
around equality would recover them. Cases E1 and E4 show posteriors collapsing to
$10^{-34}$, which no correct belief should produce from the available evidence,
and which indicates that repeated multiplicative updates compound without bound.
Capping or re-calibrating repeated-claim evidence is the direct fix. Case E5 is a
different failure entirely, since the policy violates a hard rule, which argues
for explicit penalty shaping on non-Coup choices in 10+ coin states rather than
for better probability estimates.

\section{Challenges and Limitations}
Coup is partially observable, so many states remain weakly identifiable from
public actions alone. In early and symmetric phases belief values stay close
across players, which limits confidence in any single-step recommendation, and
this is a property of the game rather than only of the implementation. The core
interactions are also multi-stage, running from action to block to challenge to
reveal, and keeping those branches consistent across replay, simulation, and
interface introduces engineering complexity and annotation noise, where a small
event-ordering error can propagate into both belief and advice.

The learning results carry the standard offline constraints. BC and DQN are
trained from logged simulator trajectories, so they are exposed to distribution
shift and extrapolation error on states the logs never covered, and policy
quality can degrade precisely in the rare, high-impact situations that decide
games. The opponents themselves are scripted styles, namely honest, bluffer,
aggressive, and cautious, which give controlled experiments but do not capture
adaptive human deception, long-horizon signalling, or meta-game adjustment across
repeated matches. The simulator also omits one real rule: a proven card is not
swapped back into the deck after a successful challenge defense, which slightly
biases downstream belief statistics.

The reported benchmark was produced under an earlier parameterisation of the
belief multipliers and the threshold rule than the one documented in
Section~\ref{sec:advisor-multiseed} and Appendix~\ref{app:belief}, so a re-run
under the current configuration is outstanding. The direction of the dominant
error mode, over-eager challenging, is unlikely to reverse, since the current
likelihood ratios are larger and therefore concentrate belief faster, but the
magnitudes should be expected to move.

The decision rule is myopic. Thresholded posterior estimates with context
adjustments are transparent, but they do not optimise long-term game value, so
retaliation risk, the signalling value of information, and endgame tempo are
unmodelled. Finally, evaluation is simulation-based and reproducible but
externally unvalidated, since there is no human-in-the-loop study, no
cross-implementation comparison, and no stratification across player counts or
table dynamics.

\section{Conclusion}
CoupVisor is a reproducible Coup research pipeline built on a single event schema
and focused on two coupled decisions: what to do each round, and when to
challenge. It converts public evidence into an explicit per-hand probability and
a threshold decision, and the per-hand conversion removes a systematic false
challenge on the opening claim. Across 500 simulated games the advisor is stable
in outcome accuracy but sensitive to opponent mix in challenge F1, and its errors
are dominated by over-eager challenges traceable to uncontrolled multiplicative
belief updates. On the learning side, both BC and DQN beat random, but which one
leads is determined by the reward definition rather than by the algorithm: shaped
short-term reward favours BC, while a win-aligned reward makes DQN the strongest
policy in the benchmark. The rule layer, including forced Coup at ten coins,
keeps regulation, advice, replay, and simulation consistent throughout.

\section{Future Work}
The most immediate work is belief calibration: post-hoc calibration such as
isotonic or temperature scaling, reliability diagrams, and confidence-aware
advisor output that states uncertainty bands instead of a bare decision. This
directly targets the dominant error mode identified above. Next is a
Thompson-sampling advisor that replaces fixed thresholds with context-specific
posteriors for both challenge and action selection, benchmarked against the
Conservative, Balanced, and Aggressive modes, together with a companion
posterior-sampling belief query that scores candidate actions under sampled full
role assignments rather than always conditioning on the posterior mean, which
places exploration at the layer where hidden-role uncertainty actually lives.

On the evaluation side, a standardised scenario suite covering Tax conflicts,
Steal block trees, Assassinate-Contessa branches, forced-Coup states, and
one-influence endgames would allow deterministic policy-level comparison
independent of rollout variance. On the training side, extending from purely
offline training to constrained self-play fine-tuning, while preserving
legality-rate and forced-Coup compliance checks during training, would address the
distribution-shift limitation, and replacing the mean-squared Bellman loss with a
soft Bellman target carrying an entropy bonus would couple the exploration
schedule to the learning signal rather than applying it only at action-selection
time. Learning opponent-type embeddings would let advisor behavior adapt to table
style instead of relying on static global thresholds. Finally, human-centred
evaluation through user studies on explanation quality, decision trust, and the
actionability of top-$k$ conditional paths for novice and experienced players is
required before any claim about practical usefulness can be considered
established.

\onecolumn
\appendices

\section{Belief Update and Challenge Rule}
\label{app:belief}
This appendix gives the code behind every belief and threshold constant quoted in
the paper.

\subsection*{A.1 Likelihood ratios}
The ratio is the probability of making a claim given that the player holds the
role, divided by the probability of making it given that they do not. An honest
claim rate near $0.85$ against a bluff rate near $0.15$ yields $5.67$. Steal
blocks are lower because either of two roles can block a Steal.

\begin{verbatim}
ACTION_LIKELIHOOD_RATIOS = {
    "Tax": ("Duke", 5.7),
    "Assassinate": ("Assassin", 5.7),
    "Steal": ("Captain", 5.7),
    "Exchange": ("Ambassador", 5.7),
}

BLOCK_LIKELIHOOD_RATIOS = {
    "Foreign Aid": [("Duke", 5.7)],
    "Assassination": [("Contessa", 5.7)],
    "Steal": [("Captain", 4.5), ("Ambassador", 4.5)],
}

CHALLENGE_WIN_RATIO = 10.0
CHALLENGE_LOSE_RATIO = 0.05
\end{verbatim}

\subsection*{A.2 Applying evidence}
Each event type multiplies the score of the role it implicates. A won challenge
is near-certain confirmation, a lost challenge near-certain disconfirmation.

\begin{verbatim}
def _apply_action_multiplier(belief, event):
    if event.actor not in belief.scores:
        return
    entry = ACTION_LIKELIHOOD_RATIOS.get(event.action_name)
    if entry is None:
        return
    role_name, ratio = entry
    role = Role(role_name)
    belief.scores[event.actor][role] *= float(ratio)


def _apply_block_multiplier(belief, event):
    if event.blocker not in belief.scores:
        return
    for role_name, ratio in BLOCK_LIKELIHOOD_RATIOS.get(event.blocked_action, []):
        role = Role(role_name)
        belief.scores[event.blocker][role] *= float(ratio)


def _apply_challenge_multiplier(belief, event):
    if event.challenged not in belief.scores:
        return
    if event.result == "win":
        scale = CHALLENGE_WIN_RATIO
    else:
        scale = CHALLENGE_LOSE_RATIO
    belief.scores[event.challenged][event.claimed_role] *= scale
\end{verbatim}

\subsection*{A.3 Deck pressure}
Every role is damped by how many of its three copies are still unrevealed, raised
to the power $0.7$ so that the damping is sublinear. A role with no copies left
is driven to zero.

\begin{verbatim}
def _apply_deck_pressure(belief, public_state):
    for role in Role:
        revealed = int(public_state.revealed_dead.get(role, 0))
        remaining = max(0, 3 - revealed)
        if remaining <= 0:
            damp = 0.0
        else:
            damp = (remaining / 3.0) ** 0.7
        for player in belief.scores:
            belief.scores[player][role] *= damp
\end{verbatim}

\subsection*{A.4 Per-hand probability}
This function implements (\ref{eq:hyper}) and (\ref{eq:blend}). The $0.7$ and
$0.3$ weights are the blend that trades combinatorial correctness against
responsiveness to accumulated claim evidence.

\begin{verbatim}
def _hypergeometric_p_truth(p_role, hand_size, remaining_copies, deck_size=15):
    from math import comb

    remaining_copies = max(0, int(remaining_copies))
    hand_size = max(1, int(hand_size))
    deck_size = max(hand_size, int(deck_size))

    if remaining_copies <= 0:
        return 0.0

    non_copies = deck_size - remaining_copies
    if non_copies < 0:
        return 1.0

    if hand_size > non_copies or hand_size > deck_size:
        p_zero = 0.0
    else:
        numerator = comb(non_copies, hand_size)
        denominator = comb(deck_size, hand_size)
        p_zero = numerator / denominator if denominator > 0 else 0.0

    p_hyper = 1.0 - p_zero
    p_independence = 1.0 - (1.0 - _clamp(p_role)) ** hand_size
    p_blended = 0.7 * p_hyper + 0.3 * p_independence

    return _clamp(p_blended)
\end{verbatim}

\subsection*{A.5 Threshold rule}
Every threshold constant quoted in the body appears here: the base values by
claim context, the style deltas, and the three table-state adjustments.

\begin{verbatim}
def _threshold_for_claim(claim_context, *, style="Balanced",
                         public_state=None, perspective=None):
    remaining = int(claim_context.remaining_copies)
    if remaining <= 0:
        base = 0.99
    elif claim_context.block_type == "Steal":
        base = 0.20
    elif claim_context.action_name == "Assassinate":
        base = 0.35
    elif claim_context.block_type == "Assassination":
        base = 0.35
    else:
        base = 0.25

    adjusted = float(base) + _style_threshold_delta(style)

    if public_state is not None:
        alive_players = [
            name for name, state in public_state.players.items()
            if int(state.influence_alive) > 0
        ]
        num_alive = len(alive_players)

        if num_alive <= 2:
            adjusted += 0.10
        elif num_alive <= 3:
            adjusted += 0.05

        if perspective is not None and perspective in public_state.players:
            my_influence = int(public_state.players[perspective].influence_alive)
            if my_influence == 1:
                adjusted -= 0.08

        for name in alive_players:
            if perspective and name == perspective:
                continue
            if int(public_state.players[name].coins) >= 7:
                adjusted += 0.06
                break
    return _clamp(adjusted, 0.01, 0.99)


def _style_threshold_delta(style):
    normalized = _normalize_style(style)
    if normalized == "Conservative":
        return -0.05
    if normalized == "Aggressive":
        return 0.07
    return 0.0
\end{verbatim}

\subsection*{A.6 Decision}
The recommendation is the comparison itself, together with the explanation string
that carries the probability, the threshold, and the inputs that produced them.

\begin{verbatim}
p_truth = _hypergeometric_p_truth(p_role_c, hand_size, remaining_copies)

threshold = _threshold_for_claim(
    claim_context, style=style, public_state=public_state, perspective=claimant,
)
if p_truth < threshold:
    recommendation = "Challenge"
else:
    recommendation = "Do not challenge"
\end{verbatim}

\section{Reward Definitions}
\label{app:reward}
The paper's central finding is that the reward definition, not the algorithm,
decided which learned policy won. This appendix gives both definitions.

\subsection*{B.1 Per-transition reward}
The shaped reward is the coin gain of the acting player plus the number of
players eliminated on that transition. Under the win-aligned mode the per-step
term is zeroed, and each transition is instead labelled with whether the acting
player went on to win the game.

\begin{verbatim}
shaped_reward = (
    (next_state["actor_coins"] - pre_coins)
    + (pre_alive - next_state["players_alive"])
)
reward = float(shaped_reward) if reward_mode == "shaped" else 0.0

...

alive_players = [
    name for name, state in engine.public_state.players.items()
    if state.influence_alive > 0
]
winner = alive_players[0] if len(alive_players) == 1 else None
for row in rows:
    row["actor_is_winner"] = int(winner is not None and row["actor"] == winner)

if reward_mode == "win":
    for row in rows:
        row["reward"] = 0.0
\end{verbatim}

\subsection*{B.2 Environment step}
The two modes differ in a single condition. Shaped mode adds $+1$ whenever the
chosen action matches the logged action, so it rewards imitating every player in
the log. Win mode adds $+1$ only when the chosen action matches the logged action
\emph{and} the player who took it went on to win, so it rewards imitating
winners only. Since the per-step term is zero in win mode, the win reward is
exactly that indicator.

\begin{verbatim}
row = self.current_rows[self.cursor]
chosen_action = self.action_labels[int(action_index) % len(self.action_labels)]
logged_action = row.get("action_name")

reward = float(row.get("reward", 0.0))
if self.reward_mode == "shaped":
    if chosen_action == logged_action:
        reward += 1.0
else:
    if chosen_action == logged_action and int(row.get("actor_is_winner", 0)) == 1:
        reward += 1.0

done = bool(int(row.get("done", 0)))
next_state = self._next_state_vector(row)
\end{verbatim}

\section{Training Setup}
\label{app:train}

\subsection*{C.1 Network}
The action-value function is a two-hidden-layer multilayer perceptron with ReLU
activations and hidden width 128.

\begin{verbatim}
class _Model(nn.Module):
    def __init__(self, in_dim, out_dim, hid_dim):
        super().__init__()
        self.net = nn.Sequential(
            nn.Linear(in_dim, hid_dim),
            nn.ReLU(),
            nn.Linear(hid_dim, hid_dim),
            nn.ReLU(),
            nn.Linear(hid_dim, out_dim),
        )

    def forward(self, x):
        return self.net(x)
\end{verbatim}

\subsection*{C.2 Defaults}
The training entry point fixes every hyperparameter quoted in the paper,
including the learning rate, the discount factor, the replay and warm-up sizes,
the target-network interval, the seed, and both exploration schedules.

\begin{verbatim}
def train_dqn_policy(
    transitions_csv,
    out_dir,
    *,
    episodes=200,
    seed=42,
    gamma=0.98,
    learning_rate=1e-3,
    batch_size=64,
    buffer_size=10000,
    warmup_steps=256,
    target_update_steps=200,
    hidden_dim=128,
    epsilon_start=1.0,
    epsilon_end=0.05,
    epsilon_decay=0.995,
    exploration_mode="epsilon_greedy",
    tau_start=1.0,
    tau_end=0.1,
    tau_decay=0.995,
    compare_bc_dir=None,
    reward_mode="shaped",
):
\end{verbatim}

\subsection*{C.3 Optimiser}
The learning rate is applied through Adam over the policy network parameters.

\begin{verbatim}
optimizer = torch.optim.Adam(policy.parameters(), lr=float(learning_rate))
\end{verbatim}

\subsection*{C.4 Action selection and entropy}
Both exploration modes are implemented in the same loop, and each computes the
per-step entropy of its own action distribution. For $\varepsilon$-greedy the
entropy follows in closed form from $\varepsilon$ and the action count, while for
the Boltzmann policy it is computed from the softmax probabilities and therefore
also reflects the spread of the learned action values. This is the measurement
plotted in Fig.~\ref{fig:exploration-modes}.

\begin{verbatim}
if exploration_mode == "epsilon_greedy":
    if rng.random() < epsilon:
        action_index = rng.randrange(action_dim)
    else:
        action_index = policy.act(state)
    p_greedy = 1.0 - epsilon + epsilon / action_dim
    p_other = epsilon / action_dim
    step_entropy = float(
        -(p_greedy * np.log(p_greedy + 1e-12)
          + (action_dim - 1) * p_other * np.log(p_other + 1e-12))
    )
else:
    with torch.no_grad():
        q_arr = policy.model(
            torch.tensor(state, dtype=torch.float32).unsqueeze(0)
        ).squeeze(0).numpy()
    scaled = q_arr / max(float(tau), 1e-6)
    scaled = scaled - float(np.max(scaled))
    probs = np.exp(scaled)
    probs = probs / probs.sum()
    action_index = int(
        rng.choices(range(action_dim), weights=probs.tolist(), k=1)[0]
    )
    step_entropy = float(-np.sum(probs * np.log(probs + 1e-12)))
\end{verbatim}

\subsection*{C.5 Replay, Bellman target, and target sync}
Updates begin only after both the batch size and the warm-up threshold are met.
The target is the standard one-step Bellman backup with the terminal flag
masking the bootstrap term, and the target network is copied from the policy
network on a fixed step interval.

\begin{verbatim}
step = env.step(action_index)
replay.push(state, action_index, step.reward, step.next_state, step.done)

if len(replay) >= int(batch_size) and len(replay) >= int(warmup_steps):
    batch = replay.sample(batch_size, rng)
    s_batch = torch.tensor([item[0] for item in batch], dtype=torch.float32)
    a_batch = torch.tensor([item[1] for item in batch], dtype=torch.long)
    r_batch = torch.tensor([item[2] for item in batch], dtype=torch.float32)
    ns_batch = torch.tensor([item[3] for item in batch], dtype=torch.float32)
    d_batch = torch.tensor([item[4] for item in batch], dtype=torch.float32)

    q_values = policy.model(s_batch).gather(1, a_batch.unsqueeze(1)).squeeze(1)
    with torch.no_grad():
        max_next_q = target.model(ns_batch).max(dim=1).values
        target_q = r_batch + float(gamma) * (1.0 - d_batch) * max_next_q

    loss = loss_fn(q_values, target_q)
    optimizer.zero_grad()
    loss.backward()
    optimizer.step()

if int(target_update_steps) > 0 and global_steps % int(target_update_steps) == 0:
    target.load_state_dict(policy.state_dict())
\end{verbatim}

\subsection*{C.6 Summary}
Table~\ref{tab:hyper} collects the values above. The exploration-mode ablation
varies only the exploration mode and its schedule, holding everything else fixed.
The behavior-cloning baseline uses a logistic-regression head over the same
feature vector, and the supervised comparison uses a Random Forest.

\begin{table}[h]
\centering
\caption{DQN training defaults.}
\label{tab:hyper}
\begin{tabular}{ll}
\toprule
Parameter & Value \\
\midrule
Network & 2 hidden layers, width 128, ReLU \\
Optimiser & Adam, learning rate $1\times10^{-3}$ \\
Loss & Mean squared Bellman error \\
Discount factor $\gamma$ & 0.98 \\
Batch size & 64 \\
Replay buffer size & 10{,}000 \\
Warm-up steps & 256 \\
Target-network update interval & 200 steps \\
Seed & 42 \\
Episodes (shaped-reward run) & 100 \\
Episodes (win-reward run) & 1000 \\
$\varepsilon$ schedule & $1.0 \rightarrow 0.05$, decay 0.995/episode \\
$\tau$ schedule (softmax mode) & $1.0 \rightarrow 0.1$, decay 0.995/episode \\
Reward modes & shaped, win-aligned \\
\bottomrule
\end{tabular}
\end{table}

\section{The Game of Coup}
\label{app:game}
This appendix summarises the published rules of Coup, so that the modelling
choices in the paper can be checked against the game they describe.

\subsection*{D.1 Setup and Goal}
The deck holds fifteen character cards, three copies each of Duke, Assassin,
Captain, Ambassador, and Contessa. The cards are shuffled and two are dealt to
each player, who may look at them but must keep them face down. The remainder
forms the Court deck in the middle of the table. Each player receives two coins
from the Treasury, and all money is kept visible. The goal is to eliminate the
influence of every other player and be the last survivor.

\subsection*{D.2 Influence}
The face-down cards in front of a player are that player's influence, and the
characters printed on them are the characters the player may legitimately claim.
Whenever a player loses an influence they turn one of their own cards face up,
choosing which one themselves. A revealed card stays face up, is visible to
everyone, and no longer provides influence. A player who has lost both cards is
exiled from the game and returns all coins to the Treasury. The game ends when
one player remains.

\subsection*{D.3 Turn Structure}
Play proceeds clockwise. On each turn the active player chooses exactly one
action and may not pass. Once the action is declared, the other players are given
an opportunity to challenge it or to counteract it, and challenges are resolved
before the action or counteraction they target. An action that is neither
challenged nor counteracted succeeds automatically. Once play moves on, a
challenge can no longer be issued retroactively.

\subsection*{D.4 Actions}
A player may choose any action they can afford. The general actions in
Table~\ref{tab:general-actions} are always available and require no role claim.
The character actions in Table~\ref{tab:character-actions} each require the
player to claim a specific character, truthfully or not. No card need be shown
unless the claim is challenged.

\begin{table}[h]
\centering
\caption{General actions, available to every player without a role claim.}
\label{tab:general-actions}
\begin{tabular}{lll}
\toprule
Action & Effect & Can be blocked by \\
\midrule
Income & Take 1 coin & Nobody \\
Foreign Aid & Take 2 coins & Duke \\
Coup & Pay 7 coins, target loses an influence & Nobody \\
\bottomrule
\end{tabular}
\end{table}

A player who begins a turn holding ten or more coins is required to launch a Coup
as their only action. A Coup always succeeds and cannot be blocked or challenged,
since it involves no role claim.

\begin{table}[h]
\centering
\caption{Character actions. Each requires claiming the listed character.}
\label{tab:character-actions}
\begin{tabular}{llll}
\toprule
Character & Action & Effect & Can be blocked by \\
\midrule
Duke & Tax & Take 3 coins & Nobody \\
Assassin & Assassinate & Pay 3 coins, target loses an influence & Contessa \\
Captain & Steal & Take 2 coins from another player & Captain, Ambassador \\
Ambassador & Exchange & Draw 2 from the Court deck, return 2 & Nobody \\
\bottomrule
\end{tabular}
\end{table}

If the target of a Steal holds only one coin, only that coin is taken. An
Exchange lets the player choose freely between their current cards and the two
drawn, returning any two cards to the Court deck afterwards.

\subsection*{D.5 Counteractions}
Counteractions block another player's action and work exactly like character
actions: the blocker claims a character, truthfully or not, need show nothing
unless challenged, and succeeds automatically if unchallenged
(Table~\ref{tab:counteractions}). When an action is successfully counteracted the
action fails, but any coins paid as its cost remain spent.

\begin{table}[h]
\centering
\caption{Counteractions.}
\label{tab:counteractions}
\begin{tabular}{lll}
\toprule
Character claimed & Blocks & Result \\
\midrule
Duke & Foreign Aid & Target receives no coins that turn \\
Contessa & Assassination & Assassination fails, fee stays spent \\
Captain or Ambassador & Steal & Thief receives no coins that turn \\
\bottomrule
\end{tabular}
\end{table}

\subsection*{D.6 Challenges}
Any action or counteraction that uses character influence can be challenged, and
any player may issue the challenge whether or not they are involved. A challenged
player must prove the claim by showing the relevant character card. If they
cannot, or choose not to, they lose the challenge. If they can, the challenger
loses it. Whoever loses immediately loses an influence.

A player who wins a challenge by showing the card returns it to the Court deck,
reshuffles, and draws a random replacement. They have therefore not lost an
influence, and the other players no longer know which card they hold. This swap
is the one published rule the simulator in this paper omits, as noted in
Challenges and Limitations. If an action is successfully challenged the entire
action fails and any coins paid as its cost are returned.

\subsection*{D.7 Double Loss on Assassination}
It is possible to lose two influence in a single turn. A player who challenges an
Assassin and loses the challenge loses one influence for the failed challenge and
then a second for the successful assassination. The same applies to a player who
bluffs a Contessa block, is challenged, and cannot show the card.

\subsection*{D.8 Relation to the Model}
The five characters are the support of the belief state. The coin costs and the
forced Coup at ten coins are the legality constraints enforced in the simulator,
in event validation, and in the interface. The blocking structure in
Tables~\ref{tab:character-actions} and~\ref{tab:counteractions} is what makes a
single turn expand into the branch structure of Fig.~\ref{fig:turn-sequence}, and
it is the reason a Steal block carries a lower challenge threshold than other
claims, since two different characters can make it. The challenge resolution rule
supplies the strongest evidence the belief tracker ever receives, which is why a
won challenge multiplies the claimed role by $10.0$ and a lost one by $0.05$.


\begin{thebibliography}{99}
\bibitem{kaelbling1998pomdp}
L.~P. Kaelbling, M.~L. Littman, and A.~R. Cassandra, ``Planning and acting in
partially observable stochastic domains,'' \emph{Artificial Intelligence},
vol.~101, no.~1--2, pp.~99--134, 1998.

\bibitem{watkins1992qlearning}
C.~J. C.~H. Watkins and P.~Dayan, ``Q-learning,'' \emph{Machine Learning},
vol.~8, no.~3--4, pp.~279--292, 1992.

\bibitem{breiman2001rf}
L.~Breiman, ``Random forests,'' \emph{Machine Learning}, vol.~45, no.~1,
pp.~5--32, 2001.

\bibitem{ross2011dagger}
S.~Ross, G.~Gordon, and D.~Bagnell, ``A reduction of imitation learning and
structured prediction to no-regret online learning,'' in \emph{Proc.\ 14th Int.\
Conf.\ Artificial Intelligence and Statistics (AISTATS)}, PMLR 15,
pp.~627--635, 2011.

\bibitem{osa2018imitation}
T.~Osa, J.~Pajarinen, G.~Neumann, J.~A. Bagnell, P.~Abbeel, and J.~Peters, ``An
algorithmic perspective on imitation learning,'' \emph{Foundations and Trends in
Robotics}, vol.~7, no.~1--2, pp.~1--179, 2018.

\bibitem{mnih2015dqn}
V.~Mnih \emph{et al.}, ``Human-level control through deep reinforcement
learning,'' \emph{Nature}, vol.~518, pp.~529--533, 2015.

\bibitem{lin1992replay}
L.-J. Lin, ``Self-improving reactive agents based on reinforcement learning,
planning and teaching,'' \emph{Machine Learning}, vol.~8, pp.~293--321, 1992.

\bibitem{levine2020offline}
S.~Levine, A.~Kumar, G.~Tucker, and J.~Fu, ``Offline reinforcement learning:
Tutorial, review, and perspectives on open problems,'' \emph{arXiv preprint
arXiv:2005.01643}, 2020.

\bibitem{fawcett2006roc}
T.~Fawcett, ``An introduction to ROC analysis,'' \emph{Pattern Recognition
Letters}, vol.~27, no.~8, pp.~861--874, 2006.

\bibitem{niculescu2005calibration}
A.~Niculescu-Mizil and R.~Caruana, ``Predicting good probabilities with
supervised learning,'' in \emph{Proc.\ 22nd Int.\ Conf.\ Machine Learning
(ICML)}, pp.~625--632, 2005.

\bibitem{sokolova2009metrics}
M.~Sokolova and G.~Lapalme, ``A systematic analysis of performance measures for
classification tasks,'' \emph{Information Processing \& Management}, vol.~45,
no.~4, pp.~427--437, 2009.
\end{thebibliography}
\end{document}